# ArchShapeNet:An Interpretable 3D-CNN Framework for Evaluating Architectural Shapes

Jun Yin[1†], Jing Zhong[1†], Pengyu Zeng[1†], Peilin Li[2], Zixuan Dai[4], Miao Zhang[1], Shuai Lu[1,3]

1 Shenzhen International Graduate School, Tsinghua University, Shenzhen, China
2 Department of Architecture National University of Singapore, College of Design and Engineering, Singapore
3 The University of Sydney, NSW, Australia
4 South Chian University of Technology, Guangzhou, China
Corresponding author, E-mail: *shuai.lu@sz.tsinghua.edu.cn*
[†]*These authors contributed equally to this work.*

## Abstract

In contemporary architectural design, the increasing complexity and diversity of design demands have made generative plugin tools indispensable for rapidly generating preliminary design concepts and exploring innovative 3D building forms. However, the quality differences between human-designed and machine-generated 3D forms remain challenging to analyze and interpret objectively. This challenge hinders the clear identification of the advantages of human-designed forms over machine-generated ones and limits the further optimization and application of generative tools in architectural design.

To address this issue:(1) We developed **ArchForms-4000**, a dataset comprising 2,000 3D forms designed by professional architects and 2,000 forms generated using the Evomass, which is a generative plugin tool. (2) We introduced **ArchShapeNet**, a 3D convolutional neural network (3D-CNN) designed for the classification and analysis of 3D form data. To align with architectural design requirements, we integrated a 3D saliency feature analysis module, which visually highlights the key form regions that the model prioritizes, offering data-driven insights for design optimization. (3) Comparative experiments demonstrate that our model surpasses professional architects in distinguishing between human-designed and machine-generated forms, achieving a classification accuracy of 94.29%, a precision of 96.2%, and a recall of 98.51%. Moreover, our approach effectively elucidates the differences between generative plugin-generated and architect-designed forms. By examining these differences, we gain deeper insights into architects' decision-making processes in form design and identify potential limitations in generative plugin tools.

This study not only highlights the distinctive advantages of human-designed forms in spatial organization, proportional harmony, and detail refinement but also provides valuable insights for enhancing generative design tools in the future.

**Highlights**
ArchForms-4000: A dataset of 2,000 human and 2,000 AI-generated 3D forms
ArchShapeNet: A 3D-CNN achieving high-accuracy form classification
3D saliency maps enhancing design interpretability
Outperforming architects in distinguishing human vs. AI-generated forms

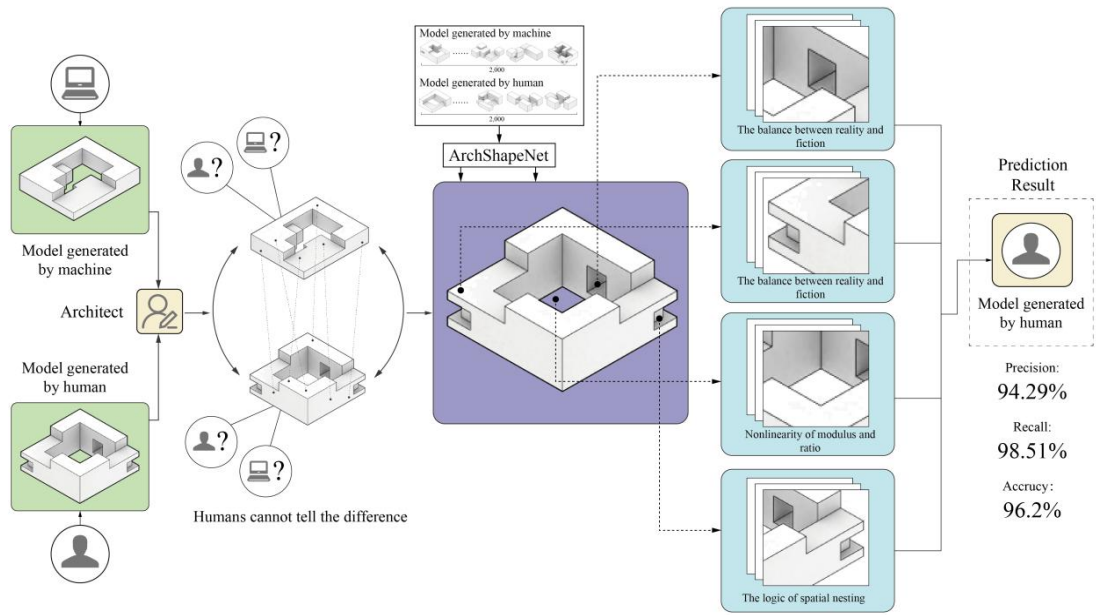

Fig. 1. Graphical Abstract

# 1. Introduction

Recent advancements in artificial intelligence (AI) have led to the widespread adoption of Deep Convolutional Neural Networks in architectural research and practice. They have significantly enhanced the efficiency of various architectural tasks, including architectural image classification (Yoshimura et al., 2019; Zhou et al., 2014), style recognition (Cantemir & Kandemir, 2024; Yi et al., 2020), scene recognition (Kong & Fan, 2020; Hu et al., 2015), and form generation (Chang, 2021). Yan et al. (2019) introduced a CNN-based method to improve the efficiency and accuracy of architectural typology classification, providing advanced tools for urban planning and geospatial analysis. Similarly, Perez et al. (2019) demonstrated the application of CNNs in architectural defect detection, enabling the automatic identification and localization of structural deficiencies. Deep learning models have also been employed in architectural style classification, with researchers investigating model interpretability to facilitate a deeper understanding of architectural style recognition and analysis (Zhang et al., 2022).

AI technologies offer designers rapid feedback and efficient iteration, thereby fostering creative potential across multiple stages of design. In particular, the rapid advancement of generative artificial intelligence, especially diffusion models and transformer-based architectures, has enabled machines to autonomously produce architecturally expressive, stylistically rich, and formally coherent outputs with minimal human input (Paananen et al., 2023; Ploennigs & Berger, 2022). Studies suggest that the visual quality and creative detailing of these outputs are, in some cases, approaching or even rivaling that of human designers (Samo & Highhouse, 2023; Chen et al., 2023). Notably, a growing number of generative AI tools are being actively integrated into mainstream design workflows. Tools like Midjourney are now widely used for architectural concept image generation and ideation (Paananen et al., 2023), TestFit has gained traction for early-stage massing and feasibility analysis in real estate development (TestFit Inc., n.d.), and EvoMass serves as a research-grade plugin for performance-driven massing studies (Wang, 2024). This accelerating integration highlights the urgent need to critically examine how AI-generated content is evaluated and distinguished within professional workflows.

Despite the increasing integration of generative AI into architectural practice, its adoption also brings notable challenges, chief among them is the black-box nature of deep learning models and the uncertainty of their outcomes, particularly in terms of interpretability (Zhou et al., 2019; Li et al., 2024). The absence of clear explanatory mechanisms prevents designers from fully comprehending the underlying logic of generated results, making it difficult to assess design rationality. Consequently, during design trade-offs and decision-making, architects are unable to leverage algorithmic insights to optimize designs for specific objectives. Additionally, many automated generative plugins rely on predefined algorithms and fixed rules, often yielding outputs that lack practical significance or fail to meet real-world design requirements. Thus, integrating interpretability techniques into deep learning models is essential. By visualizing key learned features, architects can gain a deeper understanding of model decision-making processes, enabling more informed and effective design interventions.

Among existing interpretability approaches, saliency map-based visualization techniques have become a prominent method for elucidating CNN decision-making (Simonyan et al., 2013; Selvaraju et al., 2017). This study presents ArchShapeNet, an interpretable 3D-CNN model designed to evaluate and distinguish architectural forms created by professional architects and those generated by automated design tools. Utilizing 3D saliency maps, we identify the key regions within architectural forms that the model prioritizes, correlating these features with design quality.

Experimental results indicate that ArchShapeNet achieves a classification accuracy of 94.29%, significantly exceeding the 77.51% accuracy of human classification. This improvement optimizes classification decisions and provides valuable design references. We suggest that incorporating advanced 3D-CNN techniques with interpretability tools has the potential to enhance both design process transparency and outcome reliability, which may contribute to a shift from black-box generative models to more controlled design optimization.

## 2. Related Work

### 2.1 The Application of Machine Learning in Architecture

Recent advances in machine learning have spurred a growing body of research in sustainable architecture (Zou et al., 2021), energy-aware design (Jia et al., n.d.; Zeng et al., 2025), multimodal generation (Yin et al., 2025a; Yin et al., n.d.; Zeng et al., 2025), visual enhancement (Yin et al., 2025b; Zhang et al., 2025a; Wang et al., 2025), and human–AI interaction (Zhang et al., 2024; He et al., 2025; He et al., 2024). In the architectural domain, machine learning has also been applied to design generation (Ampanavos et al., 2021), urban environment detection (Liu et al., 2017; Helber et al., 2019), and floor plan recognition (Hu et al., 2020; Jeon et al., 2023). These studies primarily focus on the analysis of 2D visual data, leveraging computer vision techniques to extract design-relevant features. For example, Pizarro et al. (2022) proposed a method for automatic floor plan classification based on 2D layout data, while Sun et al. (2022) employed street view imagery to identify building age and architectural style.

While these studies have demonstrated the potential of machine learning in architecture, they are often limited to surface-level visual analysis and lack deeper insights into spatial structures, functional layouts, and their interrelationships. Architectural design is inherently three-dimensional, involving complex interactions between form, function, and environmental adaptation. However, current machine learning models, primarily trained on 2D image datasets, struggle with higher-level design tasks such as 3D spatial layout analysis, functional zoning, and optimization of architectural form in response to environmental factors. These challenges arise due to the inherent complexity of 3D architectural models, which require models to capture spatial hierarchies, volumetric relationships, and dynamic constraints—aspects that traditional 2D-based approaches fail to address effectively.

### 2.2 The Applications of 3D-CNN

3D Convolutional Neural Networks (3D-CNNs) extend the capabilities of traditional 2D Convolutional Neural Networks (CNNs) by enabling the processing of data with three-dimensional structures. This architecture has gained widespread adoption across various domains, including medical image analysis (Çiçek et al., 2016; Singh et al., 2020; Niyas et al., 2022), video processing (Köpüklü et al., 2022; Mahadevan et al., 2020), and object detection (Hegde et al., 2016; Brazil et al., 2019), due to its ability to effectively capture spatial and temporal dependencies.

Medical imaging is one of the domains where 3D-CNNs have been applied extensively, particularly in the analysis of CT and MRI scans, where their capacity to capture complex three-dimensional spatial features can support tasks related to disease diagnosis. For instance, Kamnitsas et al. (2017) introduced an efficient multi-scale 3D-CNN for precise brain lesion segmentation, demonstrating significant improvements in accuracy. Likewise, Alakwaa, Nassef, and Badr (2017) proposed a 3D-CNN-based framework for lung cancer detection and classification, highlighting the potential of deep learning in early-stage diagnosis.

Beyond medical applications, 3D-CNNs play a pivotal role in video analysis by capturing both spatial and temporal dynamics across consecutive frames. Tran et al. (2015) developed a 3D convolutional network tailored for learning spatiotemporal features in video sequences, leading to substantial advancements in content analysis. Similarly, Haddad, Lézoray, and Hamel (2020) proposed a 3D-CNN model for facial emotion recognition from video data, enhancing accuracy by effectively modeling temporal dependencies. Such approaches have practical applications in fields like mental health, where analyzing patient video recordings can assist in diagnosing emotional disorders and behavioral conditions.

Despite their remarkable performance across diverse applications, 3D-CNNs still face critical challenges. One of the most significant limitations is their lack of interpretability, as their internal decision-making processes remain largely opaque. This black-box nature makes it difficult to assess how specific features contribute to predictions, posing a major barrier to their adoption in domains where explainability is crucial, such as architectural design and analysis. The inability to transparently interpret their outputs limits their applicability in fields requiring clear, justifiable decision-making, underscoring the need for further advancements in interpretable deep learning methodologies.

### 2.3 Research on Interpretable Maps

In recent years, the interpretability of deep learning models has gained increasing attention as a key research focus, motivated by the growing demand for transparency, accountability, and trust in artificial intelligence systems. Researchers have explored various techniques to analyze the decision-making processes of deep learning models, aiming to shed light on how these models extract and utilize features. One widely used approach involves the application of heatmaps to visualize the key regions that models focus on when making predictions, thereby offering insights into their inner workings (Zhang & Wu, 2018; Fuxin et al., 2021). Among these methods, Zhou et al. (2016) introduced Class Activation Mapping (CAM), which generates heatmaps to highlight the areas of an image that contribute most to a model's decision. This technique not only enhances interpretability but also aids in object localization by revealing the spatial attention patterns of convolutional neural networks.

Beyond general image analysis, interpretability techniques have also been applied in humanitarian efforts. In particular, target detection methods in satellite imagery have been leveraged to generate intuitive and easily interpretable poverty distribution maps, which play a crucial role in global poverty monitoring and intervention planning (Ayush, Uzkent, Burke, Lobell, & Ermon, 2020). By translating complex data into visual representations, these methods enable policymakers and researchers to make informed decisions regarding resource allocation and policy adjustments. However, while interpretability has been extensively studied in the context of 2D images, research on the interpretability of 3D data remains relatively limited.

This interpretability gap is particularly evident in the field of architectural design, where decision-making is inherently spatial, multi-scalar, and grounded in the manipulation of three-dimensional forms. Although artificial intelligence is increasingly integrated into architectural workflows, current interpretability frameworks often fall short in articulating transparent and semantically meaningful explanations for complex spatial reasoning processes. This limitation underscores the urgent need for novel interpretability techniques tailored specifically for 3D architectural data. By enabling designers to better understand how AI models interpret spatial forms and design choices, such methods could significantly enhance the practicality, reliability, and adoption of AI-driven solutions in architectural practice.

## 3. Dataset

In architectural design, there exist fundamental distinctions between architect-designed forms and those generated by computational plugins. Human-designed architecture is deeply rooted in the designer's intuition, experience, and aesthetic sensibilities, resulting in carefully curated proportions, structured mathematical modules, and intricate spatial compositions that balance voids, solids, and surface articulations. This process involves a nuanced interplay of creativity and functional considerations, allowing architects to imbue their designs with a sense of meaning, cultural context, and artistic expression. In contrast, plugin-generated forms are constructed based on predefined algorithms and parameters, enabling computational systems to rapidly produce architectural geometries in an automated manner (Wang et al., 2020). While this approach excels in handling complex forms, optimizing spatial layouts, and efficiently generating variations, it inherently lacks the subjectivity, emotional depth, and artistic refinement that characterize human-centric design.

Despite fundamental differences in design methodologies and creative processes, as illustrated in the comparative analysis in Figure 2, visually distinguishing between architect-designed and plugin-generated forms has become increasingly difficult. Recent advances in computational design and parametric modeling have enabled digital algorithms to fine-tune architectural forms with a high degree of precision, aligning them closely with design intent while accounting for structural feasibility, material constraints, and environmental considerations. These technological improvements have significantly bridged the gap between algorithmic generation and human creativity, enabling plugin-generated forms to mimic the fluidity, proportions, and spatial coherence traditionally associated with human-designed architecture. In some cases, the subtle curvature, delicate balance of proportions, and overall visual impact of computationally generated designs can be nearly indistinguishable from those crafted by experienced architects.

As a result, the once-clear boundaries between these two modes of architectural creation have begun to blur, raising critical questions about authorship, originality, and the role of human intuition in contemporary architectural practice.

To systematically investigate this increasing ambiguity, we constructed a dataset named ArchForms-4000, consisting of 2,000 architect-designed forms and 2,000 plugin-generated forms. The dataset captures a broad spectrum of spatial configurations, formal strategies, and geometric complexities that represent the characteristic differences between these two design paradigms. The development of ArchForms-4000 builds upon our prior research in architectural performance optimization (Lin et al., 2021), robust 3D modeling analysis (Lu et al., 2021), and AI-driven form generation methods (Yin et al., 2025). These works provided a rich foundation of architectural form data and a well-defined evaluation framework. Based on this foundation, we curated and integrated existing architectural shape samples and introduced two new representative categories: parametric forms generated via the EvoMass plugin; and manually modeled designs created by architects to reflect real-world creativity. This diversified composition enhances the dataset's structural coverage and control, providing a reliable basis for evaluating model interpretability and generalizability.

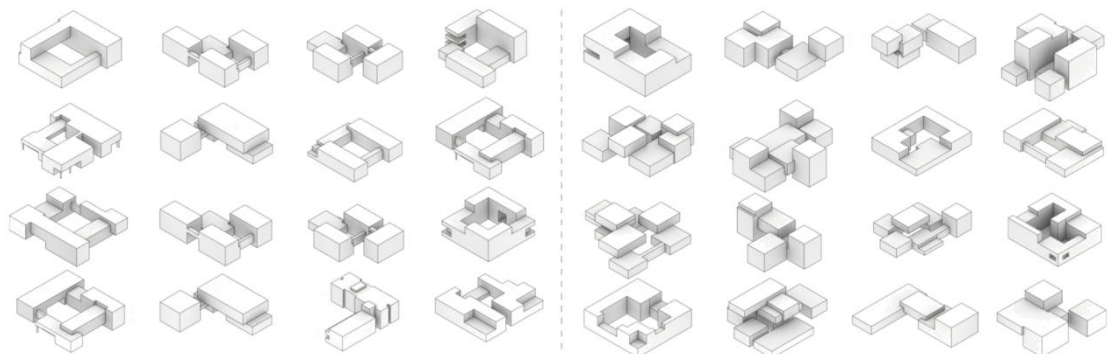

Fig. 2. left:3D forms designed by human designers. Right:3D forms generated by EvoMass

**3.1 Forms designed by EvoMass plugin**

EvoMass is an architectural design plugin for Rhino/Grasshopper based on evolutionary algorithms, focusing on the automated generation of design forms. The EvoMass plugin integrates parametric design tools with a computational design optimization approach tailored to morphological types, providing significant support for early-stage performance-based architectural form design (Wang et al., 2022a, 2024b). By combining parametric modeling with multi-objective optimization, it can efficiently generate design schemes within user-defined goals and constraints.

EvoMass was selected as the primary generative tool due to its extensive use in computational design research and its demonstrated capacity to generate formally diverse building masses through a range of parametric inputs. Specifically, EvoMass enables systematic variation of massing through parameters such as floor area ratio, geometric distortion, shear angle, and envelope articulation. This allows for the generation of a broad spectrum of forms, ranging from regular extrusions to fragmented or skewed compositions. In prior studies, EvoMass has been applied in early-stage design exploration for solar optimization, wind simulation, and life-cycle energy analysis, which affirms its suitability for producing research-grade, performance-aware architectural forms(Venus et al., 2023;

Yang et al., 2024; Shen et al., 2021; Wang & Lei, 2023).

The parameters set in EvoMass for this study are shown in Figure 3, and the corresponding 3D architectural forms generated are displayed in Figure 2 (left). In terms of parameter settings, the basic geometric features of the forms are constrained by adjusting factors such as the number of units, fill coefficients, and façade types. Additionally, interference factors are used to control the core area and the rotation-shear angle, achieving refined control over the forms, which results in shapes with various spatial orientations.

These forms include both highly regular geometric shapes and more free-form configurations generated through randomized parameters and transformation settings. This diversity ensures that the forms in the dataset exhibit high distinguishability and variability, while also simulating the complex forms that may emerge in human design.

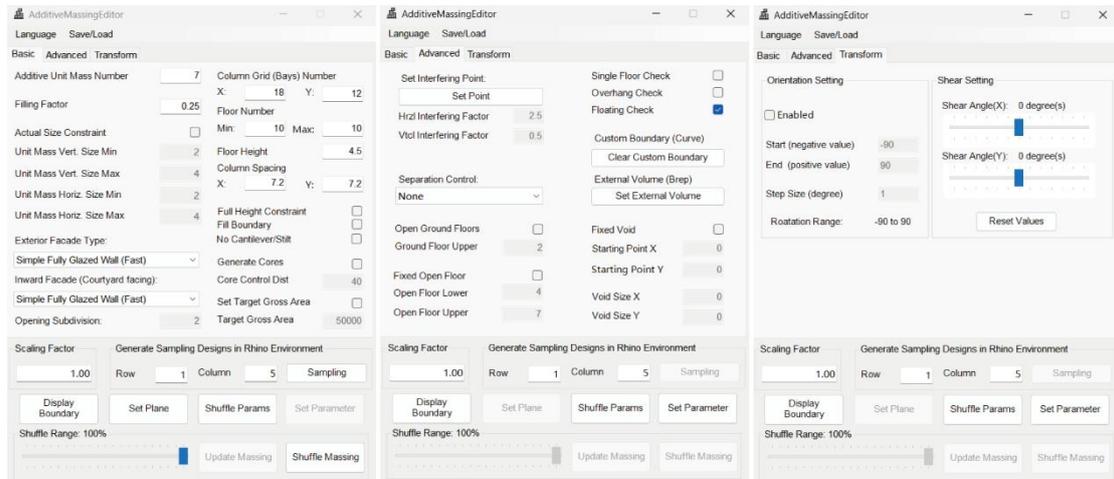

Fig. 3. EvoMass generation parameter settings

### 3.2 Forms designed by architects

Figure 2 (left) illustrates the 3D architectural forms designed by professional architects in this study. This dataset was collectively created by a total of 28 architects. To ensure that these high-quality architectural models are suitable for model training, standardization and preprocessing steps were performed.

First, non-architectural elements such as landscapes, auxiliary components, or overly complex texture details were removed to ensure data purity. Second, the scale and orientation of the models were standardized so that they adhered to a consistent scale and coordinate system, facilitating algorithmic understanding and comparative analysis. Finally, the models were simplified through voxelization or triangular mesh conversion, transforming them into 3D matrix data that can be directly processed by the 3D-CNN, thereby improving computational efficiency and optimizing the model input format.

### 3.3 Voxelization of Architectural Forms Models

All 3D architectural models used in this study were ultimately standardized into voxel representations before being input into the 3D-CNN classifier. Many of the raw models contained complex triangular mesh structures and were submitted in formats such as .obj, .3dm, and .stl. To standardize these inputs, a sequence of preprocessing steps was applied: non-architectural elements were first removed, followed by mesh simplification and then voxelization. For models with simpler geometry or cleaner mesh structures, voxelization was performed directly.

As illustrated in Figure 4, all input data were converted into regular voxel grids suitable for 3D-CNN processing. This unified representation ensures consistency in data structure and preserves key spatial features of the original architectural forms.

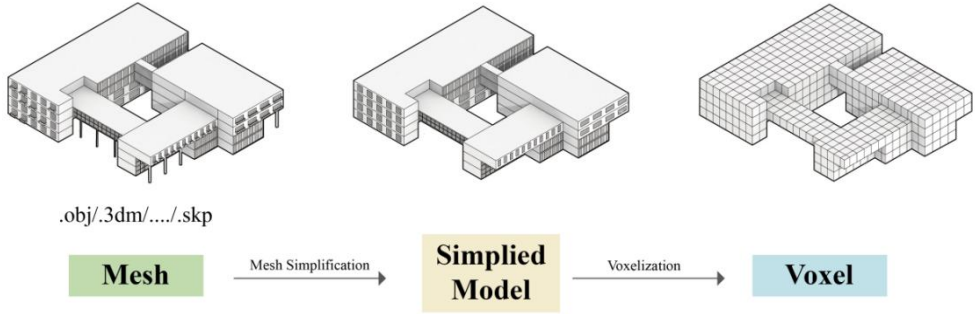

Fig. 4. Standardized preprocessing pipeline for 3D architectural models: from original mesh to voxel representation

## 4. Methodology

**4.1 3D-CNN**

The architecture of 3D-CNN refer to figure 5. We set up each convolutional layer to apply a 3D convolutional kernel (or filter) to the input data to extract spatially hierarchical features from it. Where $x \in \mathbb{R}^{C \times D \times H \times W}$ denotes the input tensor with C channels and dimensions $D \times H \times W$ (depth, height, width). $w \in \mathbb{R}^{C \times k_d \times k_h \times k_w}$ is a 3D convolution filter (convolution kernel) with dimensions $k_d \times k_h \times k_w$. $b$ is the bias term, and $y_{i,j,k}$ is the output of the convolution at spatial location $(i, j, k)$.

$$y_{i,j,k} = (x * w)_{i,j,k} + b$$

After each convolution operation, Batch Normalization (BN) is applied to normalise the output to obtain $\hat{x}$. Batch Normalization normalises each input by calculating the mean μ and variance $σ^2$ of the batch.

$$\hat{x} = \frac{x - \mu}{\sqrt{\sigma^2 + \epsilon}} \epsilon$$

$\epsilon$: a small constant used to prevent division by zero

In order to reduce the spatial dimension of the feature map and to control the computational noise, a maximum pooling operation is applied after some convolutional layers. The goal of the pooling operation is to reduce the spatial resolution by selecting the maximum value in the otherisation window.

$$y_{i,j,k} = \max_{p,q,r} x_{i+p, j+q, k+r}$$

$x$:input tensor; $y$:output tensor after pooling.

$p, q, r$ :denotes the size of the pooling window in the depth, height and width directions.

After the features are extracted in the convolution and pooling layers, the resulting feature map is spread into a vector and passed through the fully connected layer for classification, $x \in \mathbb{R}^C$ is the input vector after spreading, $W \in \mathbb{R}^{C' \times C}$ is the weight matrix of the fully connected layer, $b \in \mathbb{R}^{C'}$ is the bias vector, $y \in \mathbb{R}^{C'}$ is the output vector, and $C'$ is the number of output categories. Finally, the output is converted to category probabilities by the SoftMax function.

$$y = Wx + b$$

$$\hat{y}_i = \frac{e^{y_i}}{\sum_{j=1}^{C'} e^{y_j}}$$

$\hat{y}_i$:Probability of category i

The network is trained using Stochastic Gradient Descent (SGD) with a momentum update algorithm, where the model parameters are updated according to the gradient of the loss function on the model parameters.

$$\theta_{t+1} = \theta_t - \eta \nabla_\theta \mathcal{L}(\theta_t) + \gamma \Delta \theta_t$$

$\eta$ : Learning rate of gradient descent

Training is done using Cross-Entropy Loss, which measures the difference between the true label and the predicted probability.

$$\mathcal{L} = -\sum_{i=1}^{C'} y_i \log(\hat{y}_i)$$

$y_i$ :The true labels of class $i$ ; $\hat{y}_i$ :Predicted probability of class $i$

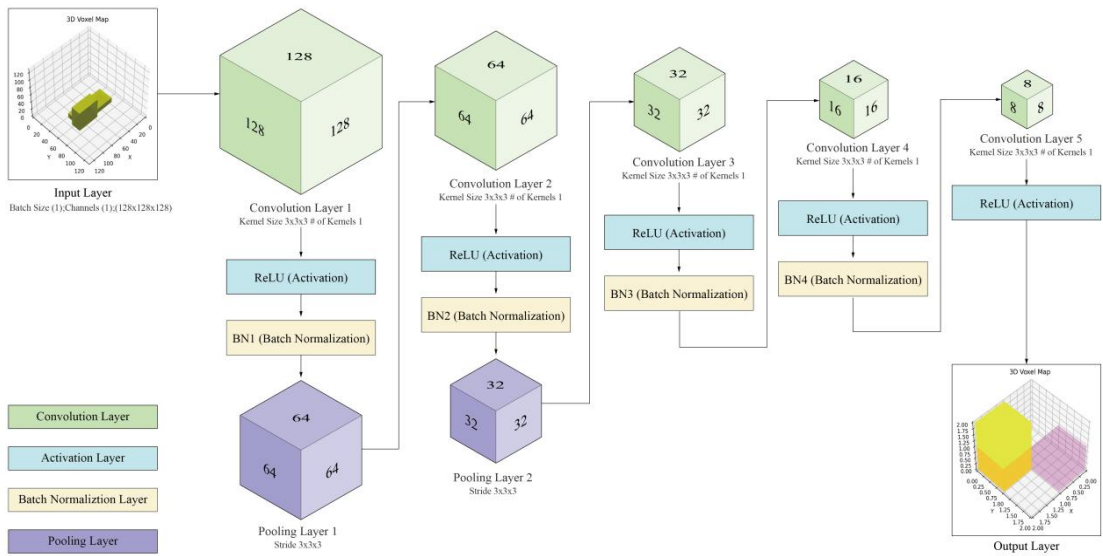

Fig. 5. 3D Convolutional Neural Network Architecture

### 4.2 3d Saliency map

The 3D voxel mesh $X \in \mathbb{R}^{128 \times 128 \times 128}$ is input to a pre-trained 3D convolutional neural network (3D-CNN), and the network output is $Y$ containing prediction scores for multiple competing categories. The score of the target category is denoted by $S_c = Y_c$, where $c$ is the target real category. The target category score $S_c$ computes a gradient about the input $X$ to obtain the effect of each voxel on the prediction result:

$$G = \frac{\partial S_c}{\partial X}$$

$G$: Each value indicates the extent to which voxels in a particular location influence the score of the target category.

An importance map of the input voxels is generated by processing the gradient. There are usually two approaches:

1. take absolute values:

$$M_{i,j,k} = |G_{i,j,k}|$$

2. square a value:

$$M_{i,j,k} = G_{i,j,k}^2$$

$M$: Importance matrix of the same size as the input $X$

In order to keep the importance values in the same scale range, the matrix $M$ is normalised and all values of Rhea are scaled to the range of $[0,1]$. The normalised matrix $\tilde{M}$ provides a normalised voxel importance map.

$$\tilde{M}_{i,j,k} = \frac{M_{i,j,k} - \min(M)}{\max(M) - \min(M)}$$

The normalisation result $\tilde{M}$ is visualised by projection or slicing. For projections, the maximum value of a shadberry axis is represented in two dimensions and the visualisation is added to the original input, showing the areas that have the most influence on the model decisions.

$$\tilde{M}_{proj}(i, j) = \max_k \tilde{M}_{i,j,k}$$

For slices, it is possible to directly extract the slice to be determined $k$:

$$\tilde{M}_{slice}(i, j) = \tilde{M}_{i,j,k}$$

## 5. Experimental Results

### 5.1 Classification Result

After completing model training, we conducted a comprehensive evaluation of the ArchShapeNet model's performance on the test set. Figure 6 illustrates the classification accuracy and overall effectiveness of the model in distinguishing between architect-designed and automatically generated architectural forms. The ArchShapeNet model achieved a classification accuracy of 94.29% and a precision of 96.2%, demonstrating its exceptional capability in differentiating between the two types of forms. Moreover, the model exhibited strong generalization ability, with no significant signs of overfitting. With a recall rate of 98.51%, it achieved high coverage of target categories, effectively identifying architectural forms while maintaining an extremely low rate of missed detections. This highlights the robustness of the model in processing and interpreting complex three-dimensional design structures.

To further compare machine classification with human evaluation, we invited a total of 16 experts, including 8 architectural PhD students and 8 experienced architects, to manually assess and classify 200 sets of 3D forms, consisting of both architect-designed shapes and those generated by EvoMass. Participants were tasked with identifying whether each form in the dataset was architect-designed or generated by EvoMass. As shown in Figure 6, our 3D-CNN model maintained closely aligned accuracy on the training and validation sets over 300 training epochs. At the same time, both training and validation loss decreased steadily without notable fluctuations or signs of overfitting. This indicates that the model did not suffer from memorization or training-set bias during feature learning, but instead acquired a generalizable understanding of geometric properties.While the model demonstrated outstanding performance across all key evaluation metrics, human classification accuracy was considerably lower, at only 77.5%. Additionally, human evaluators achieved a recall rate of 96.0% and a precision of 83.8%, indicating a tendency to misclassify certain architectural forms.

Through extensive training on large datasets, the ArchShapeNet model successfully learned to capture subtle differences in 3D shapes, maintaining consistency in its evaluation criteria while eliminating errors caused by subjective cognitive biases and fatigue. As presented in the classification data in Table 1, human evaluators primarily misclassified architect-designed forms as machine-generated ones, with 56 such misclassifications. In contrast, instances where machine-generated forms were mistaken for human-designed ones were significantly lower, with only 8 occurrences. This suggests that machine-generated architectural forms generally exhibit lower quality, making them easier for human evaluators to distinguish. However, certain architect-designed 3D forms also displayed lower complexity, particularly simple square-shaped designs. Due to their minimalistic geometric structure and lack of intricate spatial articulation, these forms were often mistaken for machine-generated shapes, contributing to classification inconsistencies.

The findings from Table 1 and Figure 7 underscore a fundamental difference in how humans and AI models approach architectural form classification. Human evaluators tend to rely heavily on intuition and prior experience, which introduces subjectivity and variability in classification accuracy. As cognitive load increases, human performance in distinguishing between the two types of forms may further decline. In contrast, the ArchShapeNet model, trained on extensive datasets, establishes a standardized and consistent evaluation framework, allowing it to maintain superior accuracy across classification tasks. These results highlight the advantages of computational models in architectural form analysis and suggest that AI-driven classification systems can play a valuable role in supplementing human expertise in architectural design evaluation.

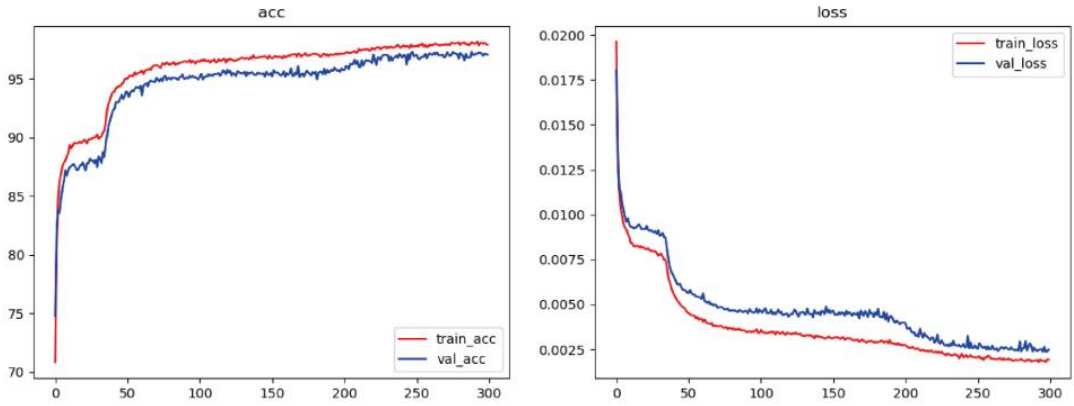

Fig. 6. ArchShapeNet model accuracy curve and loss value curve

| Evaluation subject | True Label | Predicted Label | Quantity |
|---|---|---|---|
| ArchShapeNet | Human | Human | 187 |
| | Human | Machine | 12 |
| | Machine | Human | 3 |
| | Machine | Machine | 198 |
| Human | Human | Human | 143 |
| | Human | Machine | 56 |
| | Machine | Human | 8 |
| | Machine | Machine | 193 |

Table 1. Comparison of ArchShapeNet and human evaluation results for architectural form classification

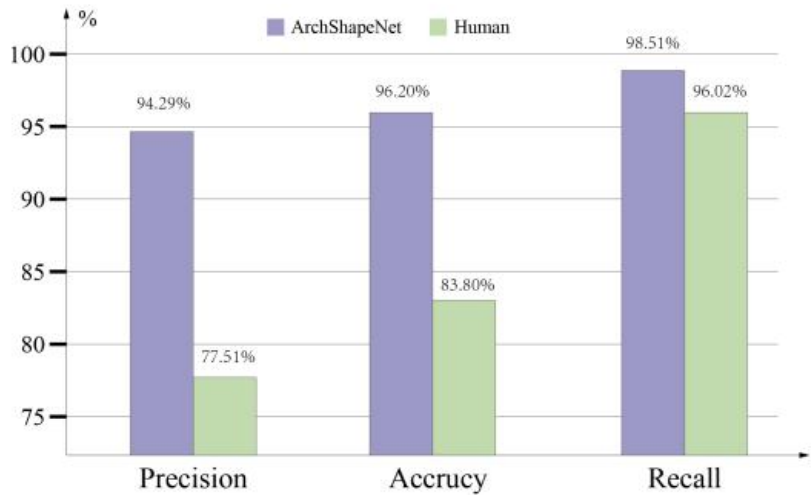

Fig. 7. Precision, accuracy, and recall metrics for ArchShapeNet and human performance.

### 5.2 Convolution result visualization

To gain deeper insights into the classification process, we employed convolution result visualization to identify the most critical regions of architectural forms that contribute to classification decisions. Specifically, we computed the gradient magnitude of the maximum predicted class score with respect to the input 3D form, allowing us to pinpoint the most sensitive voxel positions used by the model for classification. This method enabled us to analyze which spatial features were most influential in distinguishing between human-designed and machine-generated forms.

Figure 7 illustrates the convolution result visualizations, revealing the progressive feature extraction process of the ArchShapeNet model across its four convolutional layers. These layers sequentially process architectural forms, capturing increasingly abstract and complex spatial characteristics that enhance classification accuracy.

The first convolutional layer primarily focuses on capturing the basic geometric structure of the input form, emphasizing fundamental elements such as contours, edges, and overall shape. At this stage, the model identifies the foundational aspects of architectural geometry, providing an initial understanding of form boundaries. As the information propagates to the second convolutional layer, the model extracts more localized details, including edges, corners, and 3D intersections, which help define intricate spatial relationships and structural integrity. This layer enables the recognition of more nuanced design elements that distinguish different form typologies.

In the third convolutional layer, the model captures more refined spatial features, such as concave and convex variations, angular transitions, and volumetric articulation. These features are essential for distinguishing between the intentional design choices typically made by architects and the algorithmic patterns found in machine-generated forms. By the time the data reaches the fourth and final convolutional layer, the model has learned to focus on high-level, abstract geometric properties, including the overall spatial composition, symmetry, and critical design elements that characterize different architectural styles.

Through this hierarchical and multi-stage feature extraction process, ArchShapeNet effectively learns to identify significant spatial attributes and structural distinctions between human-designed and machine-generated architectural forms. By systematically analyzing the relationships between form elements at various levels of abstraction, the model gains a deep understanding of architectural composition.

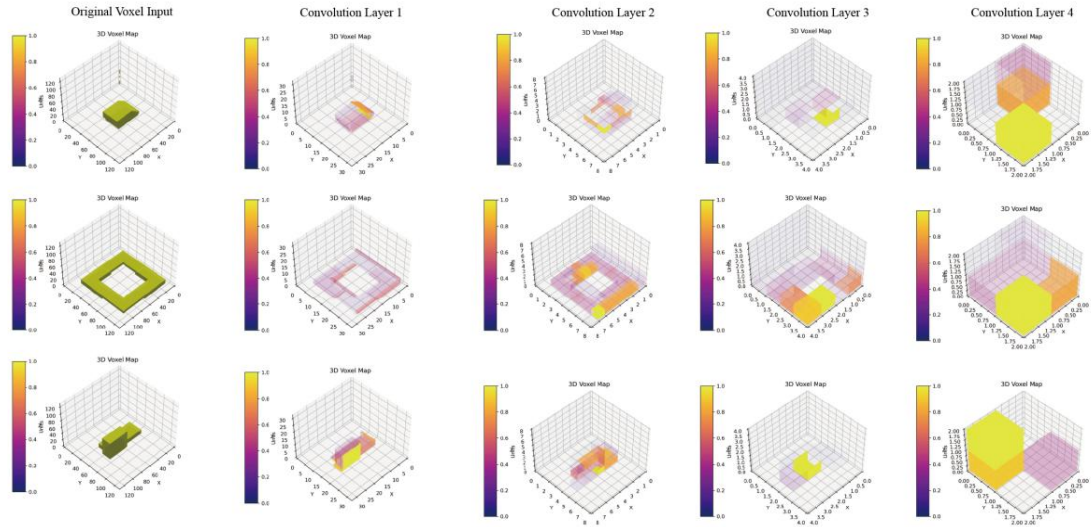

Fig. 7 Visualisation of convolution results

**5.3 Saliency Map Results Analysis**

The saliency map results for the classification task are presented in Figure 8, categorized into four distinct types: (1) human-designed forms correctly labeled as human-designed, (2) human-designed forms misclassified as machine-designed, (3) machine-generated forms correctly labeled as machine-generated, and (4) machine-generated forms misclassified as human-designed. The saliency ranking displayed in Figure 8, ranging from Rank 1 (dark blue) to Rank 10 (dark red), effectively visualizes the ArchShapeNet model's varying degrees of attention across different regions of the input models.

From the distribution of colors across the 3D forms, it is evident that the model's classification decisions are primarily influenced by two key factors: (1) Boundary features and geometric discontinuities—regions exhibiting significant geometric transitions, sharp edges, and abrupt structural changes; and (2) Complex structures and densely featured regions—areas characterized by intricate designs, high levels of detail, and elaborate geometric compositions. These focus areas suggest that the model relies heavily on local geometric properties and structural complexity when distinguishing between human-designed and machine-generated forms.

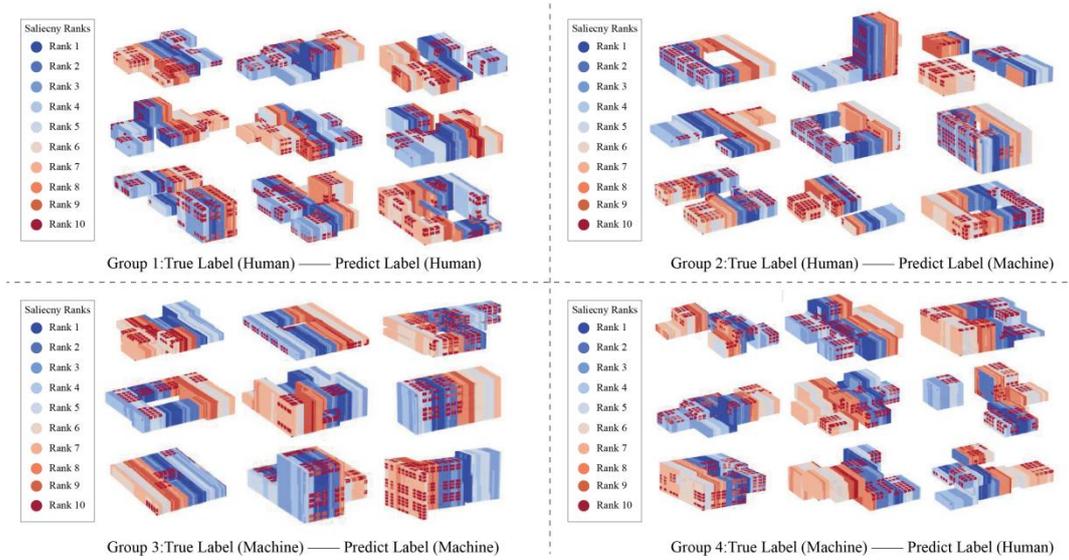

Fig. 8 Saliency map results of form classification

The model is specifically designed to interpret formal differences at the early-stage massing level, where geometric articulation and volumetric differentiation often serve as primary carriers of architectural intent. In this phase, formal decisions typically precede the assignment of detailed programmatic functions. This focus aligns with prevailing design workflows in conceptual design, where massing is treated as a vehicle for both spatial exploration and stylistic expression prior to functional resolution. By analyzing boundary operations, aggregation logic, and three-dimensional continuity, the model captures latent spatial strategies embedded in the form itself.

· **Boundary features and geometric discontinuities**

From the 3D saliency maps, the 3DCNN model assigns significantly higher attention to boundary regions and areas where geometric discontinuities occur, as these zones typically contain a higher concentration of complex geometric features that play a crucial role in shape classification. These characteristics are particularly notable in human-designed forms, which often exhibit intricate structural variations and greater heterogeneity compared to machine-generated counterparts. As illustrated in Group 1 (top left of Figure 8), the Rank 1 and Rank 2 regions, which are represented by dark blue and light blue shading, are predominantly located around sharp corners, the intersections of concave and convex surfaces, and the edges of structural frameworks. These specific regions tend to serve as the most distinctive markers for differentiating between human-designed and AI-generated architectural forms due to their higher spatial complexity and irregularity.

A similar pattern can be observed in some machine-generated forms that were misclassified as human-designed (Group 4, bottom right of Figure 8), as they also exhibit a comparable formal language, incorporating geometric intricacies that resemble the spatial complexity often found in human designs. This suggests that when AI-generated forms incorporate a certain degree of complexity, their geometric features may partially align with those of human designs, leading to potential misclassification. In contrast, plugin-generated forms that exhibit simpler geometric characteristics (Group 3, bottom left of Figure 8) tend to feature more regular shapes, symmetrical arrangements, and minimal structural articulation, particularly in these key regions. As a result, these designs often lack the intricate spatial complexity that defines human-created forms.

Given these observations, the model's capacity to effectively capture and analyze variations in boundary features, geometric discontinuities, and formal irregularities serves as a fundamental mechanism that enables it to differentiate between human-designed and machine-generated forms with a relatively high degree of accuracy. By focusing on these defining structural elements, the classification process can leverage the inherent differences in design complexity and spatial articulation to improve the robustness of the classification outcomes.

- **Complex structures and densely featured regions**

Human-designed forms often incorporate geometry driven by functional requirements, such as recesses for lighting or ventilation, which hold significant value for identification. As shown in Figure 8, Group 1 and Group 4 forms highlight Rank 3 to Rank 5 regions (blue-purple to purple-red), which cover areas with extensive detail variations, such as transition zones between structures. These regions indicate the model's focus on feature-dense parts of the form, enabling it to capture the unique variations and creativity characteristic of human design.

In contrast, plugin-generated forms (Group 3, bottom left of Figure 8) are often constrained by algorithmic procedural rules, resulting in repetitive geometric structures and a lack of organic variation. While these forms maintain a degree of coherence, their over-reliance on symmetry, uniformity, and rule-based generation mechanisms leads to a reduced diversity in complex spatial arrangements. This inherent limitation makes them less responsive to adaptive design principles, which are commonly found in human-created architectural works. The clear contrast between these two categories enables the model to systematically identify and prioritize regions with high information density, ultimately improving classification accuracy and reinforcing its ability to distinguish between functionally driven, human-inspired designs and procedurally generated architectural outputs.

- **Weaker Focus on Regular and Symmetrical Regions**

The model demonstrates significantly weaker attention to regions characterized by regularity and symmetry, as these areas tend to contain less distinctive geometric features that contribute to classification. Across all four groups analyzed, Rank 8 to Rank 10 regions (ranging from red to dark red) are predominantly located on flat surfaces, large uniform areas, or regions with minimal structural variation. These areas exhibit low information density, meaning that they provide limited discriminatory value for distinguishing between human-designed and machine-generated forms. As a result, they are generally not prioritized by the model as key reference points in the classification process.

For plugin-generated forms, which frequently incorporate highly repetitive geometric characteristics due to algorithmic constraints, the model relies on fewer high-saliency regions to correctly identify their origin. Since these forms often lack complex transitions or intricate spatial differentiations, even a small number of distinguishing features can be sufficient for classification. In contrast, while human-designed forms may also include locally symmetrical or regular structures, they are often accompanied by subtle variations, functional adaptations, or context-driven design choices that introduce a greater degree of heterogeneity. The model naturally shifts its classification focus toward these feature-dense regions, where geometric irregularities, transitions between surfaces, and localized design nuances are more apparent. By emphasizing these higher-information zones, the model enhances its ability to accurately distinguish between the two types of forms, leading to more robust classification performance.

### 5.4 Generative Plugin Optimization Suggestions Based on 3D Saliency Images

Saliency analysis validates the model's classification basis. The areas the model focuses on most during classification are concentrated on the boundaries, intersections, or regions of geometric discontinuity within the forms. The saliency in these regions directly reflects the model's reliance on specific features for classification. Conversely, the least focused areas are typically distributed across large, smooth surfaces or regions lacking variation, as these areas have lower information density.To align the results of generative plugins with the high-quality forms designed by human architects, the following optimizations should be implemented:

Enhancing Support for Complex Composite Forms: Generative plugins need to introduce greater irregularity and randomness to better mimic human preferences for complex composite spatial aesthetics. By increasing concave-convex variations, contrasts between solid and void spaces, and edge details, the generated results can better reflect the layering and aesthetic sophistication often sought in human design.

Focusing on Detailed Design in Feature-Dense Areas: Generative plugins should improve control over high-information-density regions, such as finely detailed transitions between indoor and outdoor spaces and connections between forms. These improvements can align these areas more closely with the functional and aesthetic considerations prevalent in human designs.

Optimizing the Generation of Smooth Regions: As illustrated by Rank 8 to Rank 10 (red to dark red) regions, overly simple designs in smooth areas should be minimized. For regions with insufficient variation, functional requirements can be used to introduce shape variations or decorative elements. For example, incorporating features like atriums, light wells, or other design details can enhance the practical application and aesthetic appeal of these areas, making the designs more comparable to human-designed forms.

## 6. Discussion and Future Work

While ArchShapeNet exhibits strong classification accuracy and domain-relevant interpretability on the ArchForms-4000 dataset, several limitations remain that warrant future investigation to improve the model's generalizability, architectural applicability, and explanatory depth.

(1) The current model is trained only on forms generated by the EvoMass plugin. Although EvoMass offers parametric diversity, its rule-based logic limits coverage of contemporary generative methods. We plan to incorporate tools like TestFit (Zhang et al., 2024), cellular automata, and diffusion-based frameworks (Yin et al., 2025) to cover a broader range of design logics and enhance model stability across varied generative approaches.

(2) This study focuses on geometric differences, but practical design also involves functional zoning and circulation. Future research will introduce semantic-functional data, such as space-use labels, form-function mappings, and performance strategies, to build a multi-dimensional evaluation system combining form and function.

(3) As generative AI increasingly resembles human design in style and proportion (Yin et al., 2025; Samo & Highhouse, 2023), classification becomes more challenging. We aim to build higher-fidelity datasets, including outputs from systems like ArchiDiff, to test ArchShapeNet's performance under blurred authorship. We also plan to add multimodal inputs, such as text and site context, to improve adaptability in ambiguous design settings.

(4) Although saliency maps provide useful visual cues, they lack semantic depth. Future work will integrate semantic segmentation, spatial attention, and expert annotations to develop a higher-fidelity interpretability framework, aligning with current research on bridging cognitive and computational explainability .

Overall, ArchShapeNet offers a promising foundation, but further work is needed to extend its adaptability across generative methods, integrate functional-semantic data, and refine its interpretability. These efforts will better prepare the model for real-world architectural design tasks.

## 7. Conclusions

This study explores the application of 3D convolutional neural networks (3D-CNNs) in the classification of 3D architectural forms and introduces ArchShapeNet, a model designed to differentiate between human-designed and machine-generated geometries. As artificial intelligence becomes increasingly embedded in architectural workflows, distinguishing between these two categories has emerged as a timely and relevant challenge. To assess the effectiveness of our model, we conducted a series of experiments on the ArchForms-4000 dataset, which contains a diverse spectrum of parametric architectural forms.

Our results demonstrate that ArchShapeNet achieves a classification accuracy of 94.29%, with a precision of 96.2% and a recall of 98.51%. These findings indicate that the model performs robustly in detecting underlying generative logics based on geometric features. However, given that our current training data is exclusively derived from the rule-based EvoMass plugin, the model's generalizability to alternative generative methods, such as TestFit, cellular automata, and diffusion-based frameworks, requires further investigation. Expanding the dataset to encompass a broader set of design paradigms is thus an essential step toward enhancing cross-method stability.

To improve interpretability, we incorporated saliency map visualization to identify which geometric regions most influence the model's predictions. While this approach offers a window into the internal logic of the 3D-CNN, its current implementation lacks semantic granularity. In future iterations, we plan to augment this with spatial attention mechanisms, semantic segmentation, and expert-labeled annotations to develop a more cognitively aligned interpretability framework. This could significantly bridge the gap between computational decisions and architectural reasoning. Furthermore, the current study focuses primarily on geometric analysis, yet architectural quality also stems from functional coherence and spatial logic. To that end, we propose integrating semantic-functional labels, such as space-use types and circulation paths, into future model training to support a more comprehensive form-function evaluation framework. This multi-dimensional approach could help align machine learning predictions more closely with human design judgment.

We also recognize a growing difficulty in distinguishing between human and AI-generated designs as generative AI evolves to mirror human aesthetic and formal sensibilities. To rigorously test the resilience of ArchShapeNet under these conditions, we plan to introduce high-fidelity datasets that simulate blurred authorship scenarios, such as those produced by diffusion-based design tools like ArchiDiff. Complementing this with multimodal inputs, such as site context and design intent expressed in natural language, could further enhance the model's adaptability in real-world and ambiguous design settings.

In conclusion, while ArchShapeNet demonstrates initial success in classifying 3D architectural forms and revealing interpretable patterns in design logic, it remains a foundational step. The path forward involves expanding data diversity, incorporating functional semantics, and refining interpretability strategies to meet the demands of increasingly complex and human-like generative systems. We hope this work lays a robust groundwork for future research at the intersection of architectural design, artificial intelligence, and cognitive interpretation.